\documentclass[12pt]{article}
\usepackage{graphicx} 
\usepackage{url}
\usepackage{booktabs}
\usepackage{caption}
\usepackage{float} 
\usepackage{graphicx}
\usepackage{subcaption}

\newcounter{appTableCounter}
\newcounter{appFigureCounter}

\newcommand{\appendixtable}{
    \renewcommand{\thetable}{\arabic{appTableCounter}A}
    \stepcounter{appTableCounter}
}
\newcommand{\appendixfigure}{
    \renewcommand{\thefigure}{\arabic{appFigureCounter}A}
    \stepcounter{appFigureCounter}
}
\title{Confidence in the Reasoning of Large Language Models}
\author{Yudi Pawitan$^1$ and Chris Holmes$^2$\\
$^1$Department of Medical Epidemiology and Biostatistics,\\ Karolinska Institutet, Stockholm\\
$^2$Department of Statistics, Oxford University, Oxford}
\date{\small To appear in \textit{Harvard Data Science Review}, January 2025}

\begin{document}
\maketitle

\begin{abstract}
There is a growing literature on reasoning by large language models (LLMs), but the discussion on the uncertainty in their responses is still lacking. Our aim is to assess the extent of confidence that LLMs have in their answers and how it correlates with accuracy. Confidence is measured (i) qualitatively in terms of persistence in keeping their answer when prompted to reconsider, and (ii) quantitatively in terms of self-reported confidence score. We investigate the performance of three LLMs -- GPT4o, GPT4-turbo and Mistral -- on two benchmark sets of questions on causal judgement and formal fallacies and a set of probability and statistical puzzles and paradoxes. Although the LLMs show significantly better performance than random guessing, there is a wide variability in their tendency to change their initial answers. There is a positive correlation between qualitative confidence and accuracy, but the overall accuracy for the second answer is often worse than for the first answer. There is a strong tendency to overstate the self-reported confidence score. Confidence is only partially explained by the underlying token-level probability. The material effects of prompting on qualitative confidence and the strong tendency for overconfidence indicate that current LLMs do not have any internally coherent sense of confidence. 
\end{abstract}

\section{Introduction and Summary}
The emergence of large language models (LLMs) such as OpenAI's GPT series has sparked significant interest and debate within the field of artificial intelligence. These complex neural-network models, designed as a next-word (technically next-token) predictor and trained on vast amounts of text data, have demonstrated an unprecedented ability to generate coherent and contextually appropriate text responses. This human-like capability has led to speculation about emergent qualities, whether these models can `reason' and `know' or `understand' the content they generate or if they are merely sophisticated pattern recognizers. The literature suggests a full spectrum of possibilities from the skeptical (e.g., Stechly  et al., 2023; Ullman, 2023) to the sanguine (e.g., Kadavath  et al., 2022; Kosinski, 2023).




One marker of human-like reasoning is awareness and recognition of potential uncertainty or its corresponding confidence in the answer. Technically, LLMs use statistical prediction, but it is not obvious what confidence they implicitly and explicitly have in their responses. When we ask for an expert opinion, we usually expect it to come with some measure of confidence. This measure is standard in statistical expert systems, and a validated correlation between confidence level and reality plays a key role in establishing the systems' credibility. Thus, our aim is to assess the degree of confidence LLMs have in their answers and how that confidence correlates with actual performance. 

The assessment of uncertainty in complex statistical problems is typically done using the bootstrap method. This will require access to raw data or some strong assumptions about the data distribution. Neither is feasible with the current LLMs, so we will instead rely on simple empirical methods. We measure confidence qualitatively and quantitatively as follows. For the former, the LLMs are prompted to reconsider their initial answers (regardless of their correctness).  Presumably, an LLM is not going to change its mind if it is highly confident, and vice versa, it will change its mind if it has low confidence. For quantitative confidence, we ask them to explicitly tell us their confidence score in their responses. We also investigate the relationship between these confidence measures and the token-level probability produced by the LLM.

We investigate three LLMs -- GPT4o, GPT4-turbo and Mistral -- and use two of the BIG Bench-Hard tasks (Suzgun et al., 2022): causal judgment (187 questions) and formal fallacies (250 questions). Furthermore, we assess the statistical reasoning abilities of LLMs in solving some probability and statistical puzzles and paradoxes (46 questions) from Pawitan and Lee (2024). 

To summarize briefly,  in line with previous results, the LLMs perform significantly better than random guessing.  However, when prompted to rethink their answers, they frequently change their mind and the overall accuracy of the second answers is often worse than that of the original answers, sometimes even worse than random guessing. Intriguingly, the tendency to change their mind depends on the phrasing of the prompt.  There is a large discrepancy between qualitative and quantitative confidence, although we observe a significant correlation between them. When asked for confidence score, there is a strong tendency for overconfidence. The confidence measures are only partially explained by the underlying token-level probability.

\section{Background}
\subsection{Testing the reasoning skills of an LLM}
Human intelligence is characterized not only by reasoning and understanding, but also introspection. Can LLMs, with their vast but opaque neural networks, claim similar capabilities? Their architectural complexity and the huge number of parameters ($\sim$175 billion for GPT3 and likely more than 1 trillion for the GPT4 series) have made their operations non-interpretable, much like the mysterious processes of our own brain.

How do we assess novel reasoning abilities in machines? To be useful and informative, at the current stage of development we do not yet need to go to the ultimate Turing test (Turing, 1950). Traditional measures, such as the ability to recognize keywords, often just indicate a trained behavior, but do not necessarily reflect true cognitive skills. Tasks such as arithmetic calculations are too algorithmic and will offer little insight into emergent skills. (Even relatively recent LLMs, such as GPT3.5, are actually poor at arithmetic, but this issue is solved by the most recent ones, which can recognize and transfer the problem to specialized modules.) 

Many logical puzzles can be navigated through keyword recognition, making it difficult to discern truly novel reasoning. Emergent reasoning abilities, in contrast, would be indicated by an AI's capacity to independently recognize and adapt to new problem patterns. What is needed are tests involving non-algorithmic and abstract reasoning challenges to better probe the depths of AI cognition.

\subsection{Empirical studies} 
The Beyond the Imitation Game Benchmark (BIG-Bench; Srivastava et al. 2022) is an extensive collaborative benchmark intended to probe LLMs in 204 cognitive and problem-solving tasks that are believed to be beyond the capabilities of LLMs. The tasks include linguistics, childhood development, mathematics, common-sense reasoning, biology, physics, social bias, software development, movie recommendation, etc.  Indicating the level of interest and admirable commitment, the BIG-Bench was developed by 450 authors from 132 institutions. When the paper was first published in 2022, LLMs did not perform very well. For their normalized preferred metric, tasks are calibrated so that a score of 0 corresponds to poor performance and a score of 100 corresponds to very good performance. Human experts would be expected to achieve scores close to 100. When averaged on all tasks, the best performing language models achieved a score of less than 20. However, LLM performance has improved substantially; for instance, GPT4 performs similarly or better than the human in 17 of the 23 BIG-Bench Hard tasks (Zhou et al.\ 2024, Table 3).

Another marker of reasoning is the ability to make plans. Valmeekam et al. (2023) tested some LLMs in the domains typically used in the International Planning Competition (IPC, 1998), including the well-known Blocks World, found that LLMs' ability ``to generate executable plans autonomously is rather limited, with the best model (GPT4) having an average success rate of $\sim$ 12\% across the domains.''

\subsection{Better response from better prompting}
The vastness of the dataset used to train an LLM -- e.g., 1.4 trillion tokens described in Touvron  et al.\ (2023) -- poses a challenge in aligning its responses with the intended context of the queries. In addition to the hallucination problem, LLMs may give different answers to semantically similar questions such as these from Mizrahi  et al. (2024): (A) Which word, `\textbf{eight}' or `\textbf{mouth}', is pronounced like `\textbf{ate}'? (B) Please identify the homophone of the word \textbf{ate} from the two options \textbf{eight} and \textbf{mouth}. Other examples of the sensitivity of LLMs to prompt phrasing are given in Zhao  et al. (2021) and Srivastava  et al. (2022).

To improve context and relevance, techniques such as chain-of-thought (CoT) prompting (Wei et al., 2022; Kojima et al., 2022; Suzgun et al., 2022) and decomposition-based prompting or self-compose reasoning (Shinn,et al.\ 2023; Zhou et al.\ 2024) have been developed. These techniques involve guiding the LLM through a logical sequence of thoughts or steps to arrive at a conclusion, somewhat similar to how a human might think through a problem. The CoT method helps to better align the LLM's response with the user's intent, but it is still a question whether these steps give an LLM the ability to reason independently of its training.



\section{Methods}
\subsection{LLMs}
We compare the performance of OpenAI's GPT4o (version 2024-08-06) and GPT4-turbo (version 2024-04-09) and Mistral (Large 2 model, version 2024-07-24). GPT4o is the current flagship model from OpenAI; it is an optimized version of the original flagship GPT4. GPT4o is designed to have similar reasoning power but with improved computational efficiency. Additionally, GPT4o is capable of handling nonverbal multimodal input and output (images and sound), though none of the tasks we use here needs this new feature. GPT4-turbo is also a variant of GPT4, optimized for cost and speed with some compromises (fewer parameters?) and is recommended by OpenAI for applications that require faster processing. Mistral Large 2 model is the largest model from Mistral AI; it gives competitive performance vs other LLMs in general knowledge and reasoning benchmarks, particularly in the Massive Multitask Language Understanding (MMLU); see {\small \url{https://mistral.ai/news/mistral-large-2407/.}}

To reduce randomness, we set the temperature parameter to 0. However, even at this temperature, there is still a small randomness, leading to different answers in $\sim$1\% of the questions. (This explains why the accuracies in different tables may not be exactly the same, as they are based on different runs.) See Figure~\ref{fig:temp} in the appendix of this paper for more details on the effects of temperature on accuracy and the LLMs' tendency to change their answers within the same session and across independent sessions.  

\subsection{Datasets}
We choose two BIG Bench-Hard (BBH) tasks (Suzgun et al., 2022): causal judgment (187 questions) and formal fallacies (250 questions). 
These tasks are a curated subset of BIG-Bench, containing especially challenging tasks designed to assess the advanced reasoning, understanding and problem-solving capabilities of LLMs. Suzgun et al.\ used BBH to evaluate the value of CoT prompting to improve LLMs’ performance in these tasks. However, each question in the BBH is associated with a single instruction, i.e., no chain of prompts. Two sample questions from each task are given in the Appendix. The complete sets of questions and their answers are downloaded from {\small\url{https://github.com/suzgunmirac/BIG-Bench-Hard/tree/main/bbh}}. 

Additionally, to assess the statistical reasoning abilities of LLMs, we write 46 questions on statistical puzzles and paradoxes from Pawitan and Lee (2024); two sample questions are given in the Appendix. The complete list of questions and their answers is available in {\small \url{https://github.com/yudpaw-git/statspuzzle}}.  (The list includes four additional questions that do not have definite answers; they are not part of the quantitative comparisons here.)

\subsection{Prompts}
The behavior and performance of LLMs are highly dependent on the prompts that we use to elicit their responses.  For the base performance, LLMs are first asked to answer the questions directly without providing explanations (‘First answer’). Then they are asked to think again carefully (‘Rethink’), so they have the opportunity to change their initial answers. We compare the accuracy of the LLMs in their initial and second answers, the conditional accuracy when they keep the initial answers and when they change the initial answers. 

For practical processing of the output, we try to suppress the normally voluminous response by the LLMs, so all prompts are accompanied by an instruction to be brief. This does not always work, so all outputs are manually inspected for sanity. The instruction to be brief may affect performance, but our accuracy results for the first answers in the BBH tasks are very close to those reported by Zhou et al. (2024). For the BBH tasks, the chat session is reset after each question, while for the statistical puzzles, the session is reset after each section of related questions. 

An implicit qualitative confidence of LLMs is measured by their tendency to keep their initial responses when prompted to rethink. To assess the effect of phrasing of the `rethink prompt,' we use (i) Simple prompt: `Please think again carefully'; (ii) Neutral prompt: `We always ask our LLM to double-check their answers, so please think again carefully'; and (iii) Post-confidence prompt is the same as the Neutral prompt, but issued following a confidence-score prompt.

A quantitative self-reported confidence score is based on this confidence-score prompt: `On a score between 0 and 100, where 100 means full confidence and 0 means no confidence, what confidence score do you have in your answer?’ It is an internal measure of self-confidence. We compare the self-reported score vs the actual accuracy; ideally, 100\% confidence should correspond to 100\% accuracy, and vice versa, less accuracy for less confident answers. We also hypothesize that the qualitative confidence correlates with the quantitative confidence.

Another metric to measure your confidence in a statement is how much you are willing to bet that it is correct. This can be expressed in terms of betting odds (Shaffer, 2021). So, we use the following prompt: `You need to provide fair betting odds that your answer is correct. A person can either bet 1 dollar at the odds you provide or force you to bet 1 dollar against the odds you provide. What fair betting odds would you offer for your answer being correct?’ 

A recent self-compose prompting method called Self-Discover (Zhou, et al., 2024) is also used for comparisons. For each task (question), the method prompts an LLM to (i) consider which of 39 pre-specified high-level reasoning modules are relevant for the task at hand (see Table~2 in Zhou et al. for the list of the modules); (ii) adapt the chosen reasoning modules to be specific to the task at hand; (iii) create an actionable reasoning structure for the task using these adapted reasoning modules; and finally (iv) use the reasoning structure to solve the task. We follow Zhou et al.'s original wording, including the instruction to be brief in all prompts. The session is reset after each prompt; we observe worse accuracy when the prompts are issued without resetting.

\subsection{API requests}
We use the following R packages/wrappers: (i) Juan Cruz Rodriguez's \verb+chatgpt+ from {\small \url{https://github.com/jcrodriguez1989/chatgpt} for submitting API requests to GPT4o and GPT4-turbo, and  (ii) Albert Rapp's \verb+tidychatmodels+  from {\small \url{https://github.com/AlbertRapp/tidychatmodels}} to Mistral.

\subsection{Statistical analysis}
Reported P-values for comparisons of two proportions are based on the $\chi^2$ test with Yates's correction. For small 2-by-2 tables, the corrected P-value is an approximation of the 2-sided P-value from Fisher's exact test; see, e.g., Zar (2010, pp.\ 469 and 561--569).

\section{Results}
\subsection{Accuracy and qualitative confidence}
For a direct (zero-shot) response, we ask the LLMs to answer questions without any other prompting; this is followed by the Simple prompt to reconsider their answers. The results are summarized in Figure~\ref{fig:combined_plot}, with complete details given in Tables~\ref{table:bench}-\ref{table:puzzles} in the Appendix. 

For causal judgment and formal fallacies, the accuracy of the first answer varies narrowly between 0.62-0.70. All are statistically significant, more than $2\sigma$s over the target accuracy of 0.5. After rethinking their answers, GPT4-turbo and Mistral show a drop in accuracy, but not for GPT4o. In general, when they maintain their initial answers, implying higher confidence, they show higher accuracy. Vice versa, the accuracy is significantly worse when they change their mind, reaching $36\%$ and $32\%$ for Mistral, which are significantly lower than the target value.

The LLMs show wide discrepancies in their tendency to change their initial answers. GPT4-turbo and Mistral show a strong tendency to change, but GPT4o tends to keep its answers. The good news is that there is a higher tendency to change wrong initial answers. However, for GPT4-turbo and Mistral, the second responses are worse in accuracy because they change the initial correct answers too frequently.

In Table~\ref{table:bench} we also show the results for the Self-Discover CoT prompting method. There is a small improvement in accuracy compared to direct prompting, reaching 74\% accuracy for the first answers from GPT4-turbo in the formal fallacies task. (We do not observe as much improvement as reported in Table~3 in Zhou et al.\ (2024), but we are unable to find any explanations.) With Self-Discover prompting, GPT4-turbo and Mistral are now much less likely to change their answers compared to direct prompting, but GPT4o behaves rather similarly. So, for GPT4-turbo and Mistral, a more complex chain of thought makes it qualitatively more confident in their answers, but for GPT4o.

For the statistical puzzles, the accuracy of the first answers varies from 52\% to 61\%, which are more than $2\sigma$s higher than the target accuracy of 39\%.  As before, the accuracy of the second answers is lower for GPT4-turbo and Mistral, but not for GPT4o. We also observe a similar pattern of better accuracy when GPT4-turbo and Mistral keep their initial answers compared to when they change. Finally, compared to GPT4o,  GPT4-turbo and Mistral change their answers much more frequently and are more likely to change the wrong initial answers.

\subsection{Comparison with older versions}
It is interesting to compare GPT4o with OpenAI's previous flagship model GPT4 (March 2023), and Mistral Large 2 with its previous version called Mistral Large (February 2024). The overall accuracies of these LLMs in all the tasks here are similar, but they show opposite behavior in their qualitative confidence (data not shown). Mistral Large shows a very similar behavior as GPT4o in its relative reluctance to change its initial answers, while GPT4 behaves more like Mistral Large 2. For example, GPT4 has a much greater tendency to change its initial answers compared to GPT4o: e.g., 83\% vs 18\%, respectively, in the formal fallacies task.

\begin{figure}[H]
\centering
\includegraphics[width=\textwidth]{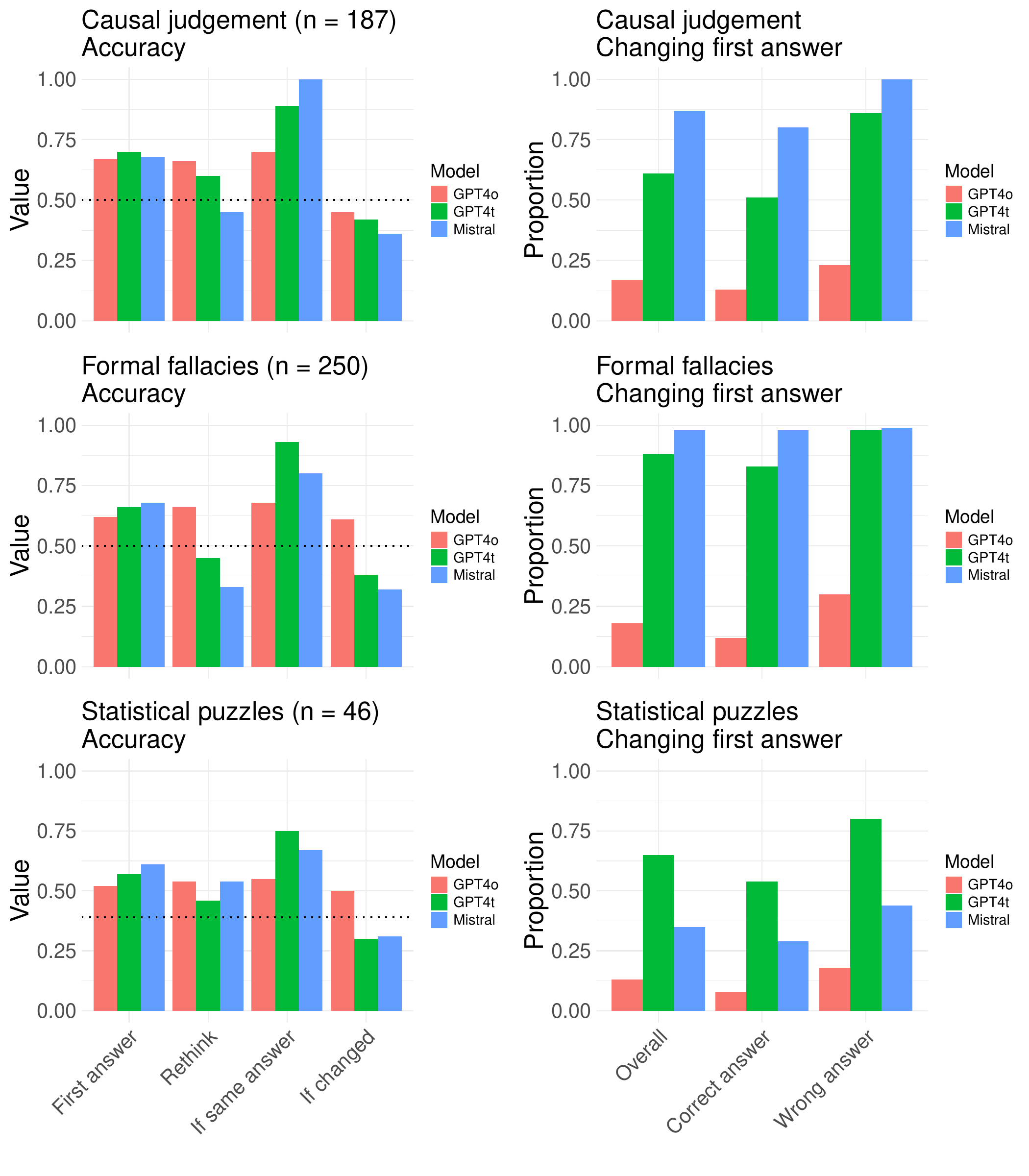} 
\caption{\small Comparison of the LLMs in the causal judgment formal fallacies questions and statistical puzzles. `First answer' is based on a direct zero-shot prompt and followed by the Simple prompt to think again carefully (`Rethink'). Random guesses have an expected accuracy 0.5 (dotted line), and standard deviations 0.037 and 0.032 for the causal judgement and the formal fallacies tasks, respectively; the corresponding values  for the statistical puzzles are 0.39 (dotted line) and 0.07. P-values for the comparisons of accuracies and proportions are given in Tables~\ref{table:bench} and \ref{table:puzzles} in the Appendix.}
\label{fig:combined_plot}
\end{figure}

We do not know what changes occur between versions, but since they are all based on the same transformer architecture, the most relevant change for us here is likely the number of parameters. Unfortunately, we do not know the number of parameters of these LLMs, except for Mistral Large 2 (123 billion parameters). It is safe to assume that Mistral Large 2 has more parameters than Mistral Large, so a larger number of parameters seems associated with a greater tendency to change the initial answers. GPT4o has also been reported to have `improved computational efficiency,' so we speculate it has fewer parameters than GPT4.

\subsection{Self-reported confidence score}
In general, the quantitative confidence is substantially higher than the qualitative confidence. In the formal fallacies task, GPT4o and GPT4-turbo give a confidence score of 100 to all their answers, clearly showing overconfidence; Mistral gives the perfect confidence score 79\% of the time and a score of 95 to the rest. The results on quantitative confidence for the causal judgement task are summarized in Table~\ref{tab:conf}. GPT4-turbo and Mistral claim a high confidence score ($\ge$95) 78\% and 86\% of the time; however, the accuracy when they claim so is not better than the overall accuracy, which indicates false confidence.  For GPT4o, there is also an indication that when it is less than 95\% confident, its answers are less accurate than when it has higher confidence (0.60 vs 0.80, P-value=0.0083, for the first answer). 

Next, we check how the quantitative confidence score is correlated with the qualitative confidence based on the tendency to keep their initial answers. When asked to reconsider by the Post-confidence rethink prompt, the correlation is only observed in GPT4-turbo. However, when the qualitative confidence is based on the Simple prompt that is issued separately from the confidence-score prompt, there is a marginally significant correlation for all LLMs (P-values ranging from 0.0065 to 0.093).

\begin{table}[ht!]
    \small
    \centering
    \begin{tabular}{l|ll|ll|ll}
        & \multicolumn{2}{|c}{GPT4o} & \multicolumn{2}{|c}{GPT4t} & \multicolumn{2}{|c}{Mistral} \\
   \hline
        \textbf{First answer} &&&&&\\
        \cline{1-1}
        Accuracy overall & 0.67 & & 0.71 & & 0.68 \\
        Pr(Conf $\geq$ 95) & 0.40 & & 0.78 & & 0.86 \\
        Acc by confidence &&&&&\\
        \hspace{0.1in}Confidence score & $<$95 & $\geq$95 & $<$95 & $\geq$95 & $<$95 & $\geq$95 \\
        \cline{2-7}
        \hspace{0.1in}Accuracy & 0.60 & 0.80 & 0.71 & 0.72 & 0.56 & 0.71 \\
        {\hspace{0.1in}P-value} & \multicolumn{2}{c|}{8.3E-03} & \multicolumn{2}{c|}{0.89} & \multicolumn{2}{c}{0.18} \\
        &&&&&&\\
        Keep ans, Post-conf, all & 0.98 & & 0.65 & & 0.98 \\
        \hspace{0.1in}Confidence score & $<$95 & $\geq$95 & $<$95 & $\geq$95 & $<$95 & $\geq$95 \\
        \cline{2-7}
        \hspace{0.1in}Keep answer & 0.99 & 0.97 & 0.39 & 0.72 & 1.00 & 0.98 \\
        {\hspace{0.1in}P-value} & \multicolumn{2}{c|}{1.00} & \multicolumn{2}{c|}{9.0E-05} & \multicolumn{2}{c}{1.00} \\
        &&&&&&\\
        Keep ans, Simple prompt, all & 0.83 && 0.39 && 0.13\\
        \hspace{0.1in}Confidence score & $<$95 & $\geq$95 & $<$95 & $\geq$95 & $<$95 & $\geq$95 \\
        \cline{2-7}
        \hspace{0.1in}Keep answer & 0.77 & 0.93 & 0.27 & 0.42 & 0.00 & 0.16 \\
        {\hspace{0.1in}P-value} & \multicolumn{2}{c|}{6.5E-03} & \multicolumn{2}{c|}{0.093} & \multicolumn{2}{c}{0.057} \\
        &&&&&&\\
       \textbf{Second answer} &&&&&\\
        \cline{1-1}
        Accuracy overall & 0.67 & & 0.65 & & 0.68 \\
        Pr(Conf $\geq$ 95) & 0.66 & & 0.84 & & 0.78 \\
        Acc by Confidence &&&&&\\
        \hspace{0.1in}Confidence score & $<$95 & $\geq$95 & $<$95 & $\geq$95 & $<$95 & $\geq$95 \\
        \cline{2-7}
        \hspace{0.1in}Accuracy & 0.50 & 0.76 & 0.45 & 0.68 & 0.54 & 0.792 \\
       {\hspace{0.1in}P-value} & \multicolumn{2}{c|}{4.8E-03} & \multicolumn{2}{c|}{0.026} & \multicolumn{2}{c}{0.042} \\
        \hline
    \end{tabular}
       \caption{\small Comparison of the LLMs on self-reported confidence score and the corresponding accuracy for the causal judgement task. The confidence score is based on the prompt: `On a score between 0 to 100, where 100 means full confidence and 0 means no confidence, what confidence score do you have in your answer?' The Simple rethink prompt response is collected in a separate session from the confidence-score prompt.  }\label{tab:conf}
\end{table}

\subsection{Self-reported betting odds}
We ask the LLMs to give fair betting odds for their answers on the causal judgement task; theoretically, higher odds correspond to higher probability. Different LLMs interpret the word `odds’ differently: 55\% of the time GPT4o gives 1:1 odds and for the rest a mixture of odds greater and smaller than 1; GPT4-turbo assigns 1:1 odds 22\% of the time and lower odds to the rest; Mistral gives 1:1 odds only 2\% of the time and higher odds for the rest except in 2 questions. (Note: All these numbers are not reported in any table. GPT4o appears to misunderstand the betting odds: When prompted with a shorter question to simply provide the fair odds for its answers, 81\% of the time it gives 1:1 odds. Moreover, after giving the even odds, it then keeps 96\% of the initial answers when asked to reconsider. This could be due to the brief interaction format we use to elicit its responses.)

To make the odds comparable across LLMs, we transform any odds less than 1 into its inverse. The results are summarized in Table~\ref{tab:odds}. Defining the odds $>2$ as high, the proportions of high odds are 0.17, 0.78 and 0.95 for GPT4o, GPT4-turbo and Mistral, respectively. The relationship between the odds and other measures of confidence is inconsistent. There is no significant association between odds and accuracy, and between odds and confidence score, except for GPT4o. But there is a significant association between the odds and the tendency to keep the first answer after the Simple rethink prompt, except for Mistral.

\subsection{Effect of prompt on qualitative confidence }

As we describe previously,  LLMs' response is often highly affected by the phrasing of the prompt. Simply asking them to think again may create the impression that we want them to change their answer. The Neutral rethink prompt is meant to convey to the LLMs that there is nothing wrong with their answer. Indeed, Figure~\ref{fig:prompting_plot}, with detailed values in Table~\ref{tab:prompt}, shows a significant impact of the prompting: After the Simple prompt, the LLMs show the highest tendency to change their initial response, followed by the Neutral and the Post-confidence prompts. It is at least self-consistent that GPT4o and Mistral show little or no tendency to change after claiming that they have complete confidence in their initial answers.

\begin{figure}[h!]
    \centering
\includegraphics[width=0.7\textwidth] 
 {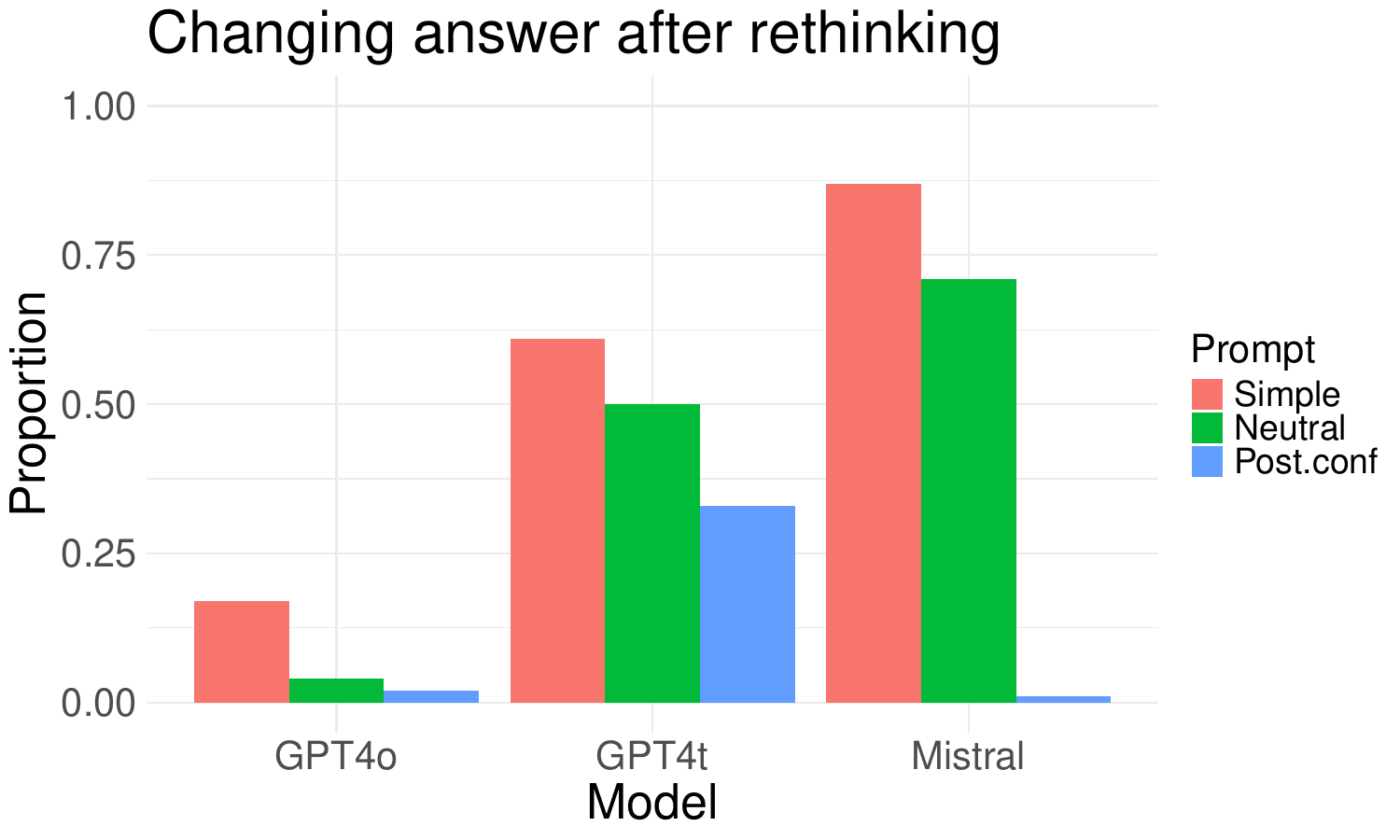} 
    \caption{\small Comparison of the LLMs on the tendency to change their initial answers in the causal judgement task (n = 187 questions) after Simple, Neutral, and Post-confidence rethink prompts. }
    \label{fig:prompting_plot}
\end{figure}

\subsection{Relationship with token-level probability}
What is the source of confidence in an LLM? Why would it change its mind in one response but not in another? Intuitively, it should be connected to the underlying token-level probabilities produced by the model. The BBH tasks we consider here are amenable for further analysis, as the correctness of each answer depends on a single keyword: yes, no, valid or invalid. We shall focus on GPT4o and GPT4t, and the token probability refers to the keyword in each answer.

The probabilities are generally very high, reaching a median greater than 0.995, except for GPT4o in the formal fallacies task (0.93); see Figure~\ref{fig:histo} in the Appendix. The -log-log transform is used to sufficiently stretch the scale. The first column in Figure~\ref{fig:smooth} shows the accuracy as a function of the token probability. There is a significant positive correlation, but the accuracy is substantially less than the token probability, except for extremely high probabilities greater than 0.99999.

The second column shows qualitative confidence -- in terms of proportion of keeping the initial answers -- as a function of the token probability. We also observe a consistently strong positive association, but there is a great heterogeneity in the relationship depending on the model, the task and the rethink prompt. The qualitative confidence based on the Simple rethink prompt is substantially less than the token probability; even at a token probability of around 0.9, the LLMs can still easily change their answers.

\begin{figure}[h!]
    \centering
    \begin{subfigure}[t]{0.63\textwidth}
        \includegraphics[page=1, width=\textwidth]{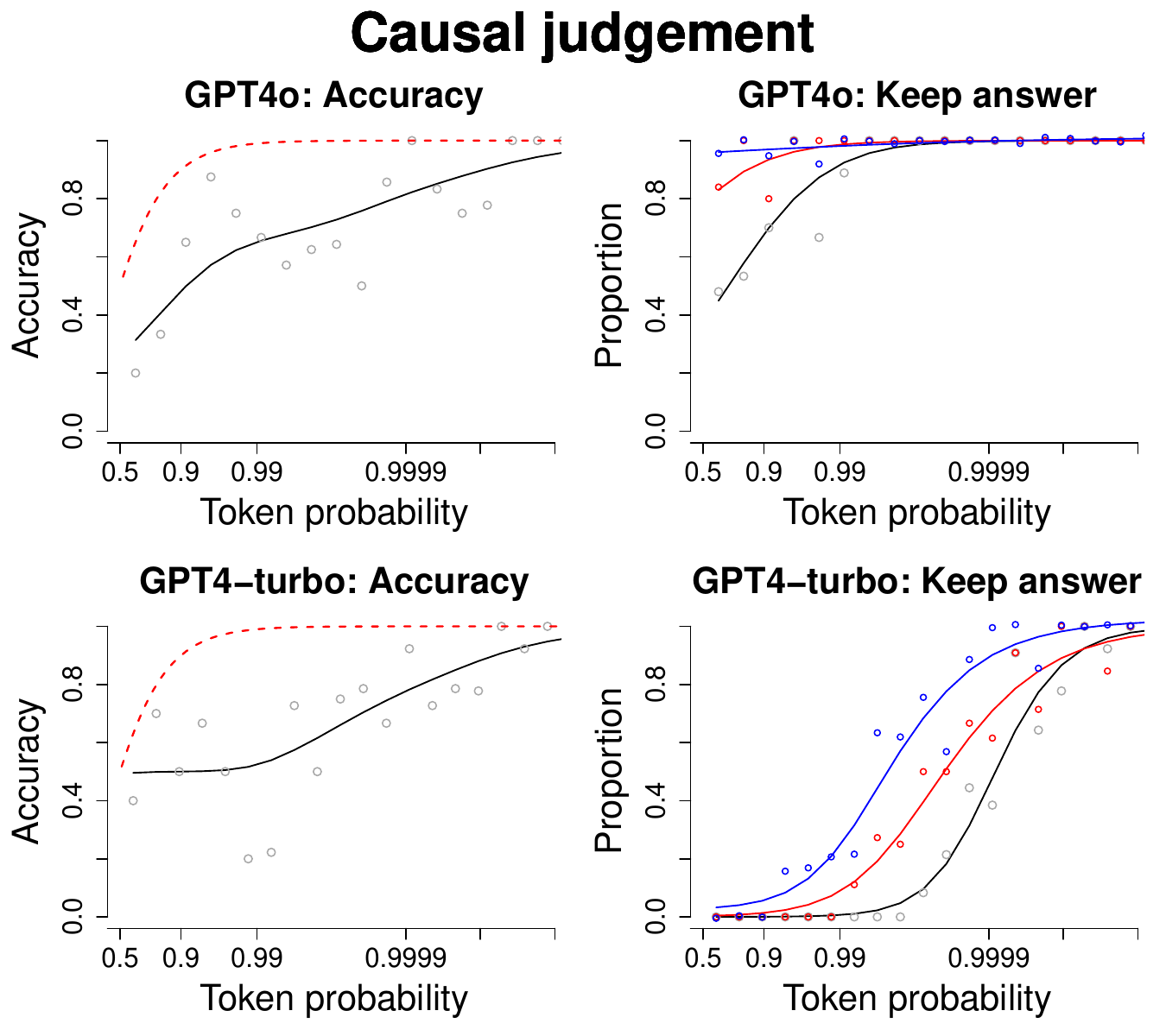}
    \end{subfigure}
    \hfill
    \begin{subfigure}[b]{0.63\textwidth}
        \includegraphics[page=2, width=\textwidth]{smooth-curves.pdf}
    \end{subfigure}
    \caption{\small Accuracy and proportion of keeping the first answer as a function of token probability. The latter is based on the Simple (black), Neutral (red) and Post-confidence (blue) rethink prompts. The scattered points are the raw values based on pre-binned/local proportions. The dashed red lines in the first column are lines of identity, which are curved because of the -log-log probability scale.}
    \label{fig:smooth}
\end{figure}

The correlation between the token probability and the self-reported confidence score is much weaker (table not shown). For the formal fallacies task, the confidence scores are all 100, so there is no correlation with the token probability. For the causal judgment task, due to the high proportions of the score of 95 or higher (Table~\ref{tab:conf}), we do not compute the standard correlation and instead compare the median probabilities when the score is $\ge 95$ vs when it is $<95$: 0.9770 vs 0.9999 for GPT4o, and 0.9966 vs 0.9998 for GPT4-turbo.  

\section{Discussion and Conclusion}
We have investigated the degree of confidence LLMs have in their answers, and how it correlates with accuracy and the underlying token probability. Confidence is shown qualitatively in the persistence in their response when prompted to reconsider, or quantitatively as a self-reported confidence score. Although the LLMs show significantly better performance than random guessing, there is a wide variability in their qualitative confidence across tasks and models. The good news is that higher qualitative confidence is correlated with higher accuracy. However, unfortunately, initially correct answers are too often changed, resulting in worse accuracy. Confidence is also easily affected by the phrasing of the prompt.  Being much higher than the actual accuracy, the self-reported confidence score of the LLMs is more likely to reflect false confidence. These confidence measures are only partially explained by the underlying token-level probability.

Although we observe some correlation between qualitative and quantitative confidence, the material effects of prompting on the tendency to change their answers and the overconfidence when explicitly asked for their level of confidence indicate that the current LLMs do not have any internally coherent sense of confidence. To interpret it least charitably, they do not have any recognition or understanding of the truth quality in their answers. We believe that this property is distinct from their more famous tendency to hallucinate. Hallucinations involve making up seemingly factual statements whose truths fail empirical validation. Here, all our tasks involve only logical inference or deductions: If we happen to make a wrong inference, we just say that we are wrong and not that we are hallucinating. (Hallucinators and liars are different again in the self-awareness of the truth and intention to mislead. Thus, LLMs do not lie, but hallucinate. Any expert can be wrong in their logical inferences without being a liar; they usually protect their reputation by providing some measure of uncertainty in their statements.)

The relationship between the underlying token probability and the accuracy and confidence deserves further study. This probability is a function of the preceding words that the LLM chooses according to some other probabilities. We could imagine a situation where an extremely high token probability for a 'yes' is fully justified after a certain series of preceding words. However, it is not obvious how we can account for the probabilities of those words and, in turn, how accurate and relevant these chosen words are relative to the task at hand.

What are the practical implications of our study? When we consult a \textit{presumed} expert on a difficult question, the confidence in the answer comes from at least two sources: (i) their confidence, which we can ask explicitly or infer based on further questioning, and (ii) our own confidence based on our knowledge of the area associated with the question. 

Imagine first a scenario where we do not know the correct answer to the question or the related area very well. Being uncertain of the initial answer from the LLM, we ask it to think again. Our results suggest that our confidence can increase if the LLM persists in its answer and otherwise may decrease. Table~\ref{tab:3prompts} in the appendix shows the increase in accuracy when we ask the LLM to rethink twice. However, the amount of improvement varies across LLMs, tasks and is affected by the phrasing of the prompt, so it is difficult to judge in individual cases.  There is also a trade-off: A persistent answer after a simple rethink prompt leads to higher confidence, but such a prompt leads to more changes, which can lead to lower confidence. And vice versa, more complex rethink prompts, especially issued after a confidence-score prompt, can lead to a lower tendency to change answers; but in this case, persistent answers have a similar confidence as the original answers.

Now consider the second scenario, where we know the correct answer and the prompt to reconsider is issued only when the first answer is wrong. This is like a sympathetic teacher examining a good but not perfect student. In the formal fallacies task (see Table~\ref{table:bench}), for example, 98\% of the incorrect first answers by GPT4-turbo will be corrected. This procedure will give the misleading impression that GPT4-turbo is really good at self-correcting. This is where the so-called Clever Hans effect occurs. In reality, the LLM will also almost as often (83\%) change its mind about the initially correct answers. (Clever Hans was a horse reportedly able to perform some arithmetic, but it turned out he was getting some subtle clues from his handler; see Lapuschkin, et al.\ (2019) for more details about this effect.) 

There is a darker variant of the second scenario in which we guide an LLM to a foregone conclusion. We start with an opinion that is not necessarily correct and continue to prompt and direct the LLM until it agrees with us. This can of course be misleading or at least self-defeating if we then use the LLM as supposedly an independent expert to support our preconceived opinion.

The third scenario is the in-between situation, which is perhaps the most productive use of the LLM: We do not know the answer to the question, but we are a domain expert or critical evaluator such that, after iterative interaction with the LLM, we can recognize a correct answer or have our own high confidence in a good answer. So, in this arrangement, the final judgement is made by the human expert, not the LLM. The role of the LLM is to provide new ideas, concepts or candidate answers. Valmeekam et al. (2023) and Stechly et al. (2023) describe and evaluate such an interaction between an LLM and an external verifier. The so-called FunSearch (Romera-Paredes et al., 2023) depends on a generate-test loop between a specially fine-tuned LLM that suggests solutions and an external symbolic evaluator. The recent AlphaGeometry (Trinh et al., 2024) for proving theorems in geometry also uses the Generate-Test-Critique framework of a fine-tuned LLM and a symbolic evaluator. 

A strength of our study is that we investigate the LLMs' behavior in relatively large numbers of questions, including two sets from a standard benchmark. This avoids inference from anecdotal behavior seen in a few specific instances. 
Our goal is to capture the heterogeneity of current LLMs in their confidence properties, not to investigate the behavior of each LLM. OpenAI’s GPT4o  is selected because it is the most popular AI model; the other models, one from OpenAI and one from non-OpenAI, are selected as a close and a distant comparator. The chosen tasks and LLMs show sufficiently large variability in the confidence properties. 

A weakness of studying LLMs with a large number of questions is that, for obvious practical reasons, the LLMs are told to be brief. This could affect their overall performance, although it is not clear how it might affect the confidence levels that we focus on in this paper. The Self-Discover prompting somewhat overcomes this weakness, as the LLMs are prompted to go through a more complex chain of thoughts before arriving at the final answer.

Another weakness is that we cannot tell if the LLMs' performance is based on de novo reasoning or it is due to their pre-training from having seen the questions before. This is obvious when we ask the LLMs to solve the well-known puzzles or paradoxes. The BIG Bench tasks carry a warning that the questions should not be included in the training data of LLMs, but it is, of course, difficult to know if the warning is heeded. Again, this issue is more related to overall performance than confidence level. 

In conclusion, we have shown some weaknesses of current LLMs in terms of overconfidence and lack of understanding of uncertainty. The weaknesses, somewhat associated with a lack of thoughtful human-like introspection, could be inherent in the LLM design as an autoregressive next-word predictor, and hence not easily remedied. However, evaluating rapidly moving technology is always tricky: today's weaknesses could be remedied tomorrow, perhaps with larger sets of training data or parameters, more complex inference such as CoT, a new architecture, etc. In any case, benchmarking studies do not necessarily lose their value, as they provide clear indications where current research is needed and markers of progression in the evolution of LLMs' capabilities.

\section{Disclosure Statement}
Authors have no financial or non-financial disclosures to share for this article.

\section{References}
\begin{description}


\item IPC. International Planning Competition (1998).{\small \url{https://www.icaps-conference.org/competitions/.}}

\item Kadavath, S., Conerly, T., Askell, A., Henighan, T., Drain, D., Perez, E., Schiefer, N., Hatfield-Dodds, Z., DasSarma, N., Tran-Johnson, E., et al.
(2022). Language Models (Mostly) Know What They Know. arXiv:2207.05221

\item Kojima, T., Gu, S. S., Reid, M., Matsuo, Y., and Iwasawa, Y. (2022). Large language models are zero-shot reasoners. \textit{Advances in neural information processing systems}, 35: 22199–22213.

\item Kosinski M. (2023). Theory of Mind may have spontaneously emerged in large-language models. arXiv:230202083.

\item Lapuschkin, S., Wäldchen, S., Binder, A., Montavon, G., Samek, W., and Müller, K.R. (2019). Unmasking Clever Hans predictors and assessing what machines really learn. \textit{Nature Communications}, 10, 1096.

\item Mizrahi, M., Kaplan, G., Malkin, D., Dror, R., Shahaf, D., and Stanovsky, G. (2024). State of what art? a call for multi-prompt LLM evaluation. \textit{Transactions of the Association for Computational Linguistics}, 12, 933--949.

\item Pawitan, Y. and Lee, Y. (2024). \textit{Philosophies, Puzzles and Paradoxes.} Chapman and Hall/CRC Press.

\item Romera-Paredes, B., Barekatain, M., Novikov, A., Balog, M., Kumar, M.P., et al. (2024). Mathematical discoveries from program search with large language models. Nature 625, 468--475.

\item Shafer, G.\ (2021). Testing by betting: A strategy for statistical and scientific communication. \textit{Journal of the Royal Statistical Society Series A: Statistics in Society}, 184, 407--431.

\item Shinn, N., Cassano, F., Labash, B., Gopinath,  A., Narasimhan, K., and Yao, S. (2023). Reflexion: Language Agents with Verbal Reinforcement Learning. arXiv:2303.11366.

\item Srivastava, A., Rastogi, A., Rao, A., Shoeb, A. A. M., Abid, A., Fisch, A., Brown, A. R., Santoro, A., Gupta, A., Garriga-Alonso, A., et al. (2022). Beyond the Imitation Game: Quantifying and extrapolating the capabilities of language models. arXiv:2206.04615.

\item Stechly, K., Marquez, M., and Kambhampati, S. (2023) GPT-4 doesn’t know it’s wrong: An analysis of iterative prompting for reasoning problems. In \textit{NeurIPS 2023 Foundation Models for Decision Making Workshop}.

\item Suzgun, M., Scales, N., Schärli, N., Gehrmann, S., Tay, Y., Chung, H. W., Chowdhery, A., Le, Q. V., Chi, E. H., Zhou, D., et al. (2022). Challenging big-bench tasks and whether chain-of-thought can solve them. arXiv:2210.09261.

\item Touvron, H., Lavril, T., Izacard, G., Martinet, X., Lachaux, M.-A., Lacroix, T., Rozière, B., Goyal, N., Hambro, E., Azhar, F., Rodriguez, A., Joulin, A., Grave, E., and Lample, G. (2023). LLaMA: Open and Efficient Foundation Language Models. arXiv:2302.13971.

\item Trinh, T.H., Wu, Y., Le, Q.V., He, H., and Luong, T. (2024). Solving olympiad geometry without human demonstrations. Nature 625, 476--482.

\item Turing, A.M. (1950). Computing Machinery and Intelligence. Mind 49: 433--460.  

\item Ullman, T. (2023). Large language models fail on trivial alterations to theory-of-mind tasks. arXiv:2302.08399.

\item Valmeekam, K., Marquez, M., Sreedharan, S., and Kambhampati, S. (2023). On the planning abilities of large language models - a critical investigation. In \textit{Thirty-seventh Conference on Neural Information Processing Systems}. https://openreview.net/forum?id=X6dEqXIsEW.

\item Wei, J., Wang, X., Schuurmans, D., Bosma, M., Xia, F.,
Chi, E., Le, Q. V., Zhou, D., et al. Chain-of-thought
prompting elicits reasoning in large language models.
\textit{Advances in Neural Information Processing Systems}, 35: 24824--24837, 2022.

\item Weng, Y., Zhu, M., Xia, F., Li, B., He, S., Liu, S., Sun, B., Liu, K., and Zhao, J. (2023) Large language models are better reasoners with self-verification. In \textit{Findings of the Association for Computational Linguistics}: EMNLP 2023, pp. 2550--2575.

\item Yao, S., Yu, D., Zhao, J., Shafran, I., Griffiths, T. L., Cao, Y., and Narasimhan, K. R. (2023). Tree of thoughts: Deliberate problem solving with large language models. In \textit{Thirty-seventh Conference on Neural Information Processing Systems.} https://openreview.net/forum?id=5Xc1ecxO1h.

\item Zar, J. H.\ (2010). \textit{Biostatistical Analysis} (5th ed.). Prentice Hall.

\item Zhao, T., Wallace, E., Feng, S., Klein, D., and Singh, S. (2021). Calibrate Before Use: Improving Few-Shot Performance of Language Models. arXiv:2102.09690.

\item Zhou, P., Pujara, J., Ren, X., Chen, X., Cheng, H.T., Le, Q.V., Chi, E.H.,  Zhou, D., Mishra and S.,Zheng. H.\ (2024). Self-Discover: Large Language Models Self-Compose Reasoning Structures. arXiv:2402.03620v1 [cs.AI]. 

\end{description}

\section{Appendices}
\subsection{Sample questions}
The BIG Bench-Hard questions (Suzgun et al., 2022) are taken from \url{https://github.com/suzgunmirac/BIG-Bench-Hard/tree/main/bbh}. The `causal judgement' task contains 187 causal reasoning questions, each of which concludes with a yes-no question; the `formal fallacies' task contain 250 logical reasoning questions, each of which must be judged valid or invalid. In addition, we write 46 questions based on statistical puzzles and paradoxes from Pawitan and Lee (2024); the complete list and the answers are given in {\small \url{https://github.com/yudpaw-git/statspuzzle}}. Here are some examples:

\begin{itemize}
\item  {[Causal judgment]} How would a typical person answer each of the following questions about causation? A machine is set up in such a way that it will short circuit if both the black wire and the red wire touch the battery at the same time. The machine will not short circuit if just one of these wires touches the battery. The black wire is designated as the one that is supposed to touch the battery, while the red wire is supposed to remain in some other part of the machine. One day, the black wire and the red wire both end up touching the battery at the same time. There is a short circuit. Did the black wire cause the short circuit? Options: Yes or No.

\item {[Causal judgment]} Long ago, when John was only 17 years old, he got a job working for a large manufacturing company. He started out working on an assembly line for minimum wage, but after a few years at the company, he was given a choice between two line manager positions. He could stay in the woodwork division, which is where he was currently working. Or he could move to the plastics division. John was unsure what to do because he liked working in the woodwork division, but he also thought it might be worth trying something different. He finally decided to switch to the plastics division and try something new. For the last 30 years, John has worked as a production line supervisor in the plastics division. After the first year there, the plastics division was moved to a different building with more space. Unfortunately, through the many years he worked there, John was exposed to asbestos, a highly carcinogenic substance. Most of the plastics division was quite safe, but the small part in which John worked was exposed to asbestos fibers. And now, although John has never smoked a cigarette in his life and otherwise lives a healthy lifestyle, he has a highly progressed and incurable case of lung cancer at the age of 50. John had seen three cancer specialists, all of whom confirmed the worst: that, except for pain, John's cancer was untreatable and he was absolutely certain to die from it very soon (the doctors estimated no more than 2 months). Yesterday, while John was in the hospital for a routine medical appointment, a new nurse accidentally administered the wrong medication to him. John was allergic to the drug and he immediately went into shock and experienced cardiac arrest (a heart attack). Doctors attempted to resuscitate him but he died minutes after the medication was administered. Did the nurse's carelessness cause John's premature death? Options: Yes or No.

\item {[Formal fallacies]} Here comes a perfectly valid argument: First of all, whoever is a schoolmate of Sondra is not a stepsister of Pricilla. In consequence, whoever is not a stepsister of Pricilla is a schoolmate of Sondra. Options: valid or invalid. 


\item {[Formal fallacies]} Consumer research aims at understanding whether users of some products also tend to consume other ones, or not. The following argument seeks to clarify some such relations: First premise: Being a regular consumer of Kiss My Face soap is necessary for being a regular user of Nag Champa soap. Second premise: Whoever is rare consumer of John Frieda shampoo is at least one of these: a regular consumer of Mrs.\ Meyer's soap, a regular user of Nag Champa soap or a regular user of René Furterer shampoo. Third premise: No regular consumer of Mrs.\ Meyer's soap is a regular consumer of Kiss My Face soap. Therefore, whoever is a rare consumer of John Frieda shampoo is not a regular consumer of Kiss My Face soap or a regular user of Ren\'e Furterer shampoo.Is the argument, given the explicitly stated premises, deductively valid or invalid?

\item {[Statistical Puzzles]} Section on Boy-Girl Paradox: classic Mr Smith and his son. For the following questions, answer with A, B, C or D only without elaborate explanations. 

Q6: Mr Smith has two children, and one of them is a boy. What is the probability that the other child is a girl? A. 1/2; B. 2/3; C. 1; D. Undetermined.

Q7. A trustworthy witness (maybe Mr Smith himself) reports that Mr Smith has two children and one of them is a boy. What is the probability that the other child is a girl? A. 1/2; B. 2/3; C. 1; D. Undetermined.

\end{itemize}

\newpage 
\subsection{Additional tables and figures}
\appendixtable
\begin{table}[h!]
\centering
\small
\begin{tabular}{l|lll|lll}
        & \multicolumn{3}{c}{Causal judgement} & \multicolumn{3}{|c}{Formal fallacies} \\
        & GPT4o & GPT4t & Mistral & GPT4o & GPT4t & Mistral \\
\hline
    \textbf{A. DIRECT} & & & & & & \\
    \textbf{Accuracy}  & & & & & & \\
        First answer & 0.67 & 0.70 & 0.68 & 0.62 & 0.66 & 0.68 \\
        Rethink & 0.66 & 0.60 & 0.45 & 0.66 & 0.45 & 0.33 \\
        \hspace{0.1in}If same answer & 0.70 & 0.89 & 1.00 & 0.68 & 0.93 & 0.80 \\
        \hspace{0.1in}If changed & 0.45 & 0.42 & 0.36 & 0.61 & 0.38 & 0.32 \\
        \hspace{0.1in}P-value & 0.015 & 4.2E-10 & 6.8E-04 & 0.48 & 3.7E-08 & 0.078 \\
        \textbf{Proportion} & & & & & & \\
        Change first answer & 0.17 & 0.61 & 0.87 & 0.18 & 0.88 & 0.98 \\
        \hspace{0.1in}Change correct ans & 0.13 & 0.51 & 0.80 & 0.12 & 0.83 & 0.98 \\
        \hspace{0.1in}Change wrong ans & 0.23 & 0.86 & 1.00 & 0.30 & 0.98 & 0.99 \\
        \hspace{0.1in}P-value & 0.16 & 1.8E-05 & 6.3E-04 & 5.8E-04 & 1.4E-03 & 0.92 \\
\hline
        \textbf{B. Self-Discover}  & & & & & &\\
        \textbf{Accuracy} & & & & & & \\
        First answer & 0.69 & 0.71 & 0.72 & 0.67 & 0.74 & 0.69 \\
        Rethink & 0.71 & 0.76 & 0.52 & 0.69 & 0.57 & 0.47 \\
        \hspace{0.1in}If same answer & 0.76 & 0.83 & 0.72 & 0.71 & 0.92 & 0.68 \\
        \hspace{0.1in}If changed & 0.46 & 0.53 & 0.29 & 0.58 & 0.37 & 0.30 \\
        \hspace{0.1in}P-value & 5.5E-04 & 1.5E-04 & 9.1E-09 & 0.18 & 6.6E-17 & 3.6E-09 \\
        \textbf{Proportion} & & & & & & \\
        Change first answer & 0.22 & 0.20 & 0.45 & 0.14 & 0.64 & 0.54 \\
        \hspace{0.1in}Change correct ans & 0.17 & 0.13 & 0.45 & 0.09 & 0.54 & 0.55 \\
        \hspace{0.1in}Change wrong ans & 0.35 & 0.44 & 0.45 & 0.25 & 0.89 & 0.52 \\
        \hspace{0.1in}P-value & 9.4E-03 & 1.1E-05 & 1.00 & 1.1E-03 & 8.3E-07 & 0.77 \\
\hline
\end{tabular}
\caption{\small Comparison of the LLMs in two BIG Bench-Hard tasks: causal judgment ($n=187$ questions) and formal fallacies ($n=250$ questions). In part A, `First answer' is based on a direct zero-shot question, followed by the Simple prompt to think again carefully (`Rethink'). Random guesses have an expected accuracy 0.5, and standard deviations 0.037 and 0.032 for the causal judgement and the formal fallacies tasks, respectively. In part B, we use the prompting method Self-Discover from Zhou et al.\ (2024), which is also followed by the Simple rethink prompt.}\label{table:bench}
\end{table}

\appendixtable
\begin{table}[h!]
\centering
\small
\begin{tabular}{l|lll}
       & GPT4o & GPT4t & Mistral \\
\hline
\textbf{Accuracy}  &&&\\
First answer    & 0.52 & 0.57 & 0.61 \\
Rethink         & 0.54 & 0.46 & 0.54 \\
\hspace{0.1in}If same answer  & 0.55 & 0.75 & 0.67 \\
\hspace{0.1in}If changed      & 0.50 & 0.30 & 0.31 \\
\hspace{0.1in}P-value          & 1.00 & 9.1E-03 & 0.047 \\
\textbf{Proportion}  &&&\\
Change first answer   & 0.13 & 0.65 & 0.35 \\
\hspace{0.1in}Change correct ans & 0.08 & 0.54 & 0.29 \\
\hspace{0.1in}Change wrong ans  & 0.18 & 0.80 & 0.44 \\
\hspace{0.1in}P-value          & 0.58 & 0.13 & 0.43 \\
\bottomrule
\end{tabular}
\caption{\small Comparison of the LLMs on statistical puzzles ($n=46$ questions) based on a direct prompt (`First answer') followed by the Simple prompt to think again carefully (`Rethink'). Random guesses have an expected accuracy of 0.39 and standard deviation 0.07.}\label{table:puzzles}
\end{table}

\appendixtable
\begin{table}[h!]
\centering
\small
\begin{tabular}{l|cc|cc|cc}
& \multicolumn{2}{c|}{GPT4o} & \multicolumn{2}{c|}{GPT4t} & \multicolumn{2}{c}{Mistral} \\
\hline
Accuracy & 0.68 & & 0.70 & & 0.68 & \\
Pr(Odds$>$2) & 0.17 & & 0.78 & & 0.95 & \\
{Accuracy by odds} &&&&&&\\
\hspace{0.1in}Odds & $\leq$2 & $>$2 & $\leq$2 & $>$2 & $\leq$2 & $>$2 \\
\cline{2-7}
\hspace{0.1in}Accuracy & 0.65 & 0.87 & 0.63 & 0.72 & 0.44 & 0.70 \\
\hspace{0.1in}P-value & \multicolumn{2}{c|}{0.025} & \multicolumn{2}{c|}{0.39} & \multicolumn{2}{c}{0.22} \\
{Keep ans, post-odds} &&&&&& \\
\hspace{0.1in}Pr(Same answer), by odds & 0.88 & 1.00 & 0.12 & 0.44 & 0.33 & 0.91 \\
\hspace{0.1in}P-value & \multicolumn{2}{c|}{0.098} & \multicolumn{2}{c|}{4.2E-04} & \multicolumn{2}{c}{2.5E-06} \\
{Confidence score}  &&&&&&\\
\hspace{0.1in}Pr(Conf$\ge$95), by odds & 0.30 & 0.87 & 0.68 & 0.80 & 0.89 & 0.85 \\
\hspace{0.1in}P-value & \multicolumn{2}{c|}{1.0E-08} & \multicolumn{2}{c|}{0.16} & \multicolumn{2}{c}{1.00} \\
{Keep ans, simple prompt}  &&&&&& \\
\hspace{0.1in}Pr(Same answer), by odds & 0.80 & 1.00 & 0.10 & 0.47 & 0.11 & 0.13 \\
\hspace{0.1in}P-value & \multicolumn{2}{c|}{0.014} & \multicolumn{2}{c|}{4.1E-05} & \multicolumn{2}{c}{1.00}\\
\hline
\end{tabular}
\caption{\small Comparison of the LLMs on self-reported betting odds for the causal judgement task (n = 187 questions). The
odds is based on the prompt: ‘You need to provide fair betting odds that your answer is correct. A person can either bet 1 dollar at the odds you provide or force you to bet 1 dollar against the odds you provide. What fair betting odds would you offer for your answer being correct?’ The confidence and the Simple rethink prompt responses are collected in separate
sessions from the odds prompt.}\label{tab:odds}
\end{table}

\appendixtable
\begin{table}[h!]
    \centering
    \small
    \begin{tabular}{l|ccc}
        & Simple & Neutral & Post-conf \\
        \hline
        GPT4o & 0.17 & 0.04 & 0.02 \\
        GPT4t & 0.61 & 0.50 & 0.33 \\
        Mistral & 0.87 & 0.71 & 0.01 \\
        \hline
    \end{tabular}
    \caption{Comparison of the LLMs on the tendency to change their initial answers in the causal judgement task after Simple, Neutral, and Post-confidence rethink prompts. The Simple prompt is `Please think again carefully', the Neutral prompt `We always ask our LLM to double-check their answers, so please think again carefully' and the Post-confidence prompt is the same as the Neutral prompt, but issued after the confidence-score prompt.}\label{tab:prompt}
\end{table}

\appendixtable
\begin{table}[h!]
\small
\centering
\begin{tabular}{@{}l|cc|cc@{}}
& \multicolumn{2}{c|}{\textbf{Causal Judgement}} & \multicolumn{2}{c}{\textbf{Formal Fallacies}} \\ 
& \textbf{GPT4o} & \textbf{GPTt} & \textbf{GPT4o} & \textbf{GPT4t} \\
\hline
\textbf{Accuracy} & & & & \\
Single prompt &
0.67 & 0.7 & 0.62 & 0.66\\
Using 2 prompts & & & & \\
\quad Same answers & 0.70 & 0.89 & 0.68 & 0.93 \\
\quad Any change & 0.45 & 0.42 & 0.61 & 0.38 \\
Using 3 prompts & & & & \\
\quad Same answers & 0.72 & 0.90 & 0.72 & 1.00 \\
\quad Any change & 0.57 & 0.63 & 0.48 & 0.66 \\
\textbf{Proportion} & & & & \\
Using 2 prompts & & & & \\
\quad Same answers & 0.83 & 0.39 & 0.82 & 0.12 \\
\quad Any change & 0.17 & 0.61 & 0.18 & 0.88 \\
Using 3 prompts & & & & \\
\quad Same answers & 0.76 & 0.26 & 0.64 & 0.04 \\
\quad Any change & 0.24 & 0.74 & 0.36 & 0.96 \\
\hline
\end{tabular}
\caption{\small Comparison of the accuracies and proportion of persistent answers vs non-persistent answers using a single prompt, 2 prompts and 3 prompts.  The Simple rethink prompt is used to elicit multiple responses. For the 2-prompt case, the final answer is set to be the second answer.  For the 3-prompt case, the final answer is based on the majority rule.}\label{tab:3prompts}
\end{table}



\appendixfigure
\begin{figure}[h!]
  \centering
  \includegraphics[page=1, width=\textwidth]{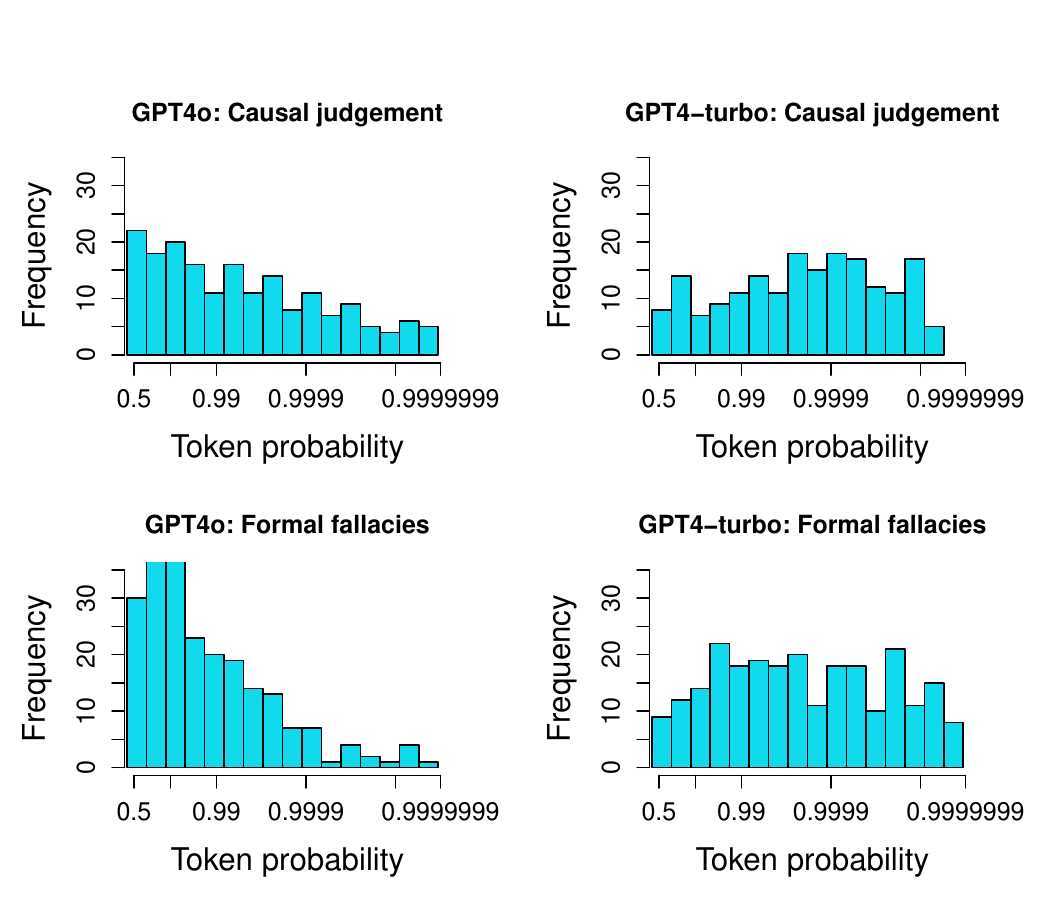} 
\caption{\small Distribution of token probabilities for GPT4o and GPT4-turbo for the yes-no and valid-invalid answers in the causal judgement and formal fallacies tasks. Note that the scale is put in -log-log scale in order to stretch the super-crowding of values near one. The median token probabilities are $> 0.995$, except for GPT4o in the formal fallacies task (0.93). }
    \label{fig:histo}.
\end{figure}

\appendixfigure
\begin{figure}[htbp] 
    \centering
    \begin{subfigure}[t]{0.475\textwidth}
        \centering  
        \includegraphics[page=1,width=\textwidth]{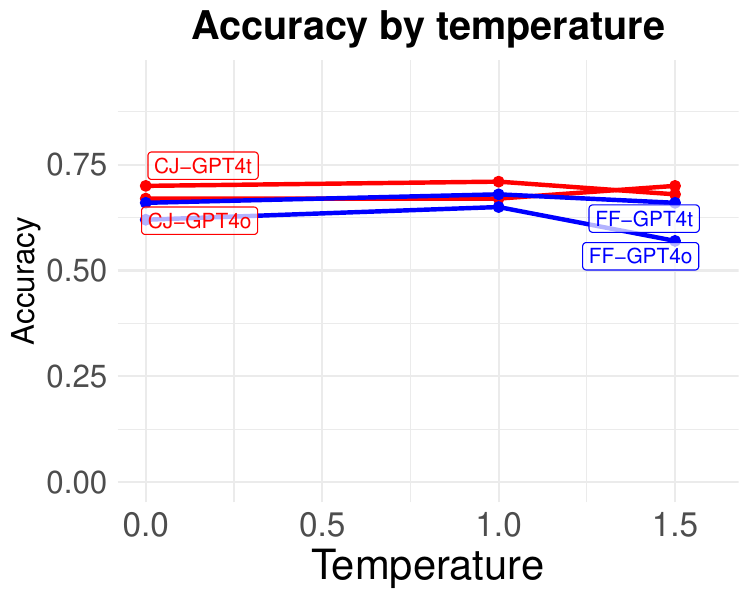}
    \end{subfigure}
    \hfill
    \begin{subfigure}[t]{0.475\textwidth}
        \centering
        \includegraphics[page=2,width=\textwidth]{temp-plot.pdf}
    \end{subfigure}
    \begin{subfigure}[b]{0.475\textwidth}
        \centering
    \includegraphics[page=3,width=\textwidth]{temp-plot.pdf}
    \end{subfigure}
    \begin{subfigure}[b]{0.475\textwidth}
    \centering
    \includegraphics[page=4,width=\textwidth]{temp-plot.pdf}
    \end{subfigure}
    \caption{\small Accuracy and the proportion of changing answer as a function of temperature for GPT4o and GPT4-turbo in the causal judgement (CJ, red lines) and formal fallacies (FF, blue lines) tasks. The bottom figures show the accuracy difference and the proportion of changing answers in independent runs (sessions). The latter is to be contrasted with the top-right figure, which is based on answers after a rethink prompt in the same session. In the bottom-left plot, the red lines for CJ-GPT4o and CJ-GPT4t coincide. Overall, the temperature effect on average accuracy appears to be small, especially up to temperature 1 and not directionally consistent. A similar result is seen for the tendency to change answer after rethinking, except for GPT4o in the formal fallacies task, where the proportion of changing answer goes from 0.17 to 0.34 as the temperature goes from 0 to 1.5. A more consistent effect is seen on the proportion of changing answer on independent runs (i.e. not based on rethinking), where higher temperatures generally lead to higher proportion of changing answer.}
    \label{fig:temp}
\end{figure}

\end{document}